\documentclass[conference, letterpaper, 10 pt]{IEEEtran}
\IEEEoverridecommandlockouts
\usepackage{cite}
\usepackage{amsmath,amssymb,amsfonts}
\usepackage{algorithmic}
\usepackage{graphicx}
\usepackage{textcomp}
\usepackage{multirow}
\usepackage[letterpaper,top=54pt,bottom=54pt,left=54pt,right=54pt]{geometry}
\usepackage{float}
\usepackage{url}
\usepackage{graphicx}
\usepackage{subcaption} 

\usepackage[table,xcdraw]{xcolor}
\def\BibTeX{{\rm B\kern-.05em{\sc i\kern-.025em b}\kern-.08em
    T\kern-.1667em\lower.7ex\hbox{E}\kern-.125emX}}

\newcommand{\dataseturl}{\url{https://syndra.retis.santannapisa.it}}
\newcommand{\name}{SynDRA-BBox}
\newcommand{\originalname}{SynDRA}

\begin{document}

\title{\vspace*{10pt}Towards Railway Domain Adaptation for LiDAR-based 3D Detection: Road-to-Rail and Sim-to-Real via \name}

\author{
\IEEEauthorblockN{
    Xavier Diaz$^1$, Gianluca D'Amico$^2$, Raul Dominguez-Sanchez$^1$, Federico Nesti$^2$, \\Max Ronecker$^1$, Giorgio Buttazzo$^2$
}\\
\IEEEauthorblockA{
    \begin{tabular}{c c}
        $^1$ Mobility Technologies, & $^2$ Department of Excellence in Robotics \& AI, \\
        SETLabs Research GmbH, Munich, Germany & Scuola Superiore Sant'Anna, Pisa, Italy \\
        $<$name$>$.$<$surname$>$@setlabs.de & $<$name$>$.$<$surname$>$@santannapisa.it
    \end{tabular}
}
}

\maketitle

\begin{abstract}
In recent years, interest in automatic train operations has significantly increased.
To enable advanced functionalities, robust vision-based algorithms are essential for perceiving and understanding the surrounding environment.
However, the railway sector suffers from a lack of publicly available real-world annotated datasets, making it challenging to test and validate new perception solutions in this domain.
To address this gap, we introduce \name, a synthetic dataset designed to support object detection and other vision-based tasks in realistic railway scenarios.
To the best of our knowledge, is the first synthetic dataset specifically tailored for 2D and 3D object detection in the railway domain, the dataset is publicly available at \dataseturl.
In the presented evaluation, a state-of-the-art semi-supervised domain adaptation method, originally developed for automotive perception, is adapted to the railway context, enabling the transferability of synthetic data to 3D object detection.
Experimental results demonstrate promising performance, highlighting the effectiveness of synthetic datasets and domain adaptation techniques in advancing perception capabilities for railway environments.

\end{abstract}


\section{Introduction}\label{intro}
\begin{figure}[H]
\centering
\includegraphics[width=0.44\textwidth]{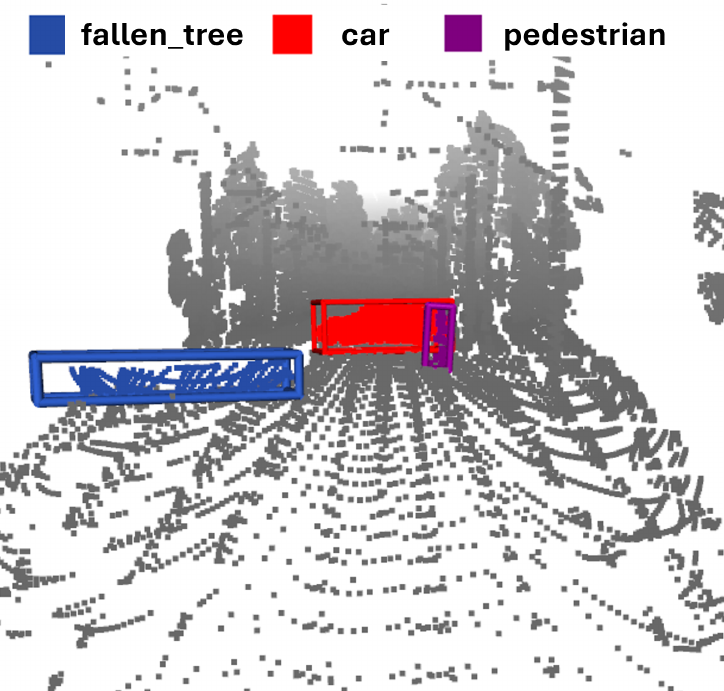}
\caption{Synthetic point cloud sample from \name. Points are colored in grey-scale based on their Z-values, while relevant object targets are colored according to their semantic class colors and enclosed within their corresponding 3D bounding boxes.}
\label{fig:point_cloud}
\end{figure}

In recent years, the railway industry has increasingly invested in research efforts to achieve higher levels of automation. 
According to the IEC 62267 standard~\cite{iec62267}, the Grade of Automation (GoA) defines the degree to which train operations are automated, ranging from GoA0, where all functions are manually performed by the driver, to GoA4, which represents a fully autonomous operation. 
While most railway systems in Europe currently operate up to GoA2, moving to GoA3 or GoA4 requires the integration of robust and accurate perception systems that comply with strict safety and reliability requirements of railway standards. 
Vision-based tasks such as semantic segmentation and 2D/3D object detection are essential to build such perception capabilities (Figure~\ref{fig:point_cloud} illustrates a synthetic point cloud from the \name~dataset, showcasing 3D bounding box annotations for different object classes.).
Unlike the automotive domain, gathering labeled multi-sensor data in the railway sector is challenging, due to safety and data protection regulations, as well as the high cost and time demands associated with data acquisition and manual labeling.
As a result, the railway domain lacks publicly available labeled datasets to train, validate, and benchmark new perception algorithms in this field.
This data scarcity significantly slows down the development and advancement of research in railway automation.
\begin{figure*}[!htb]
\centering
\includegraphics[width=0.85\textwidth]{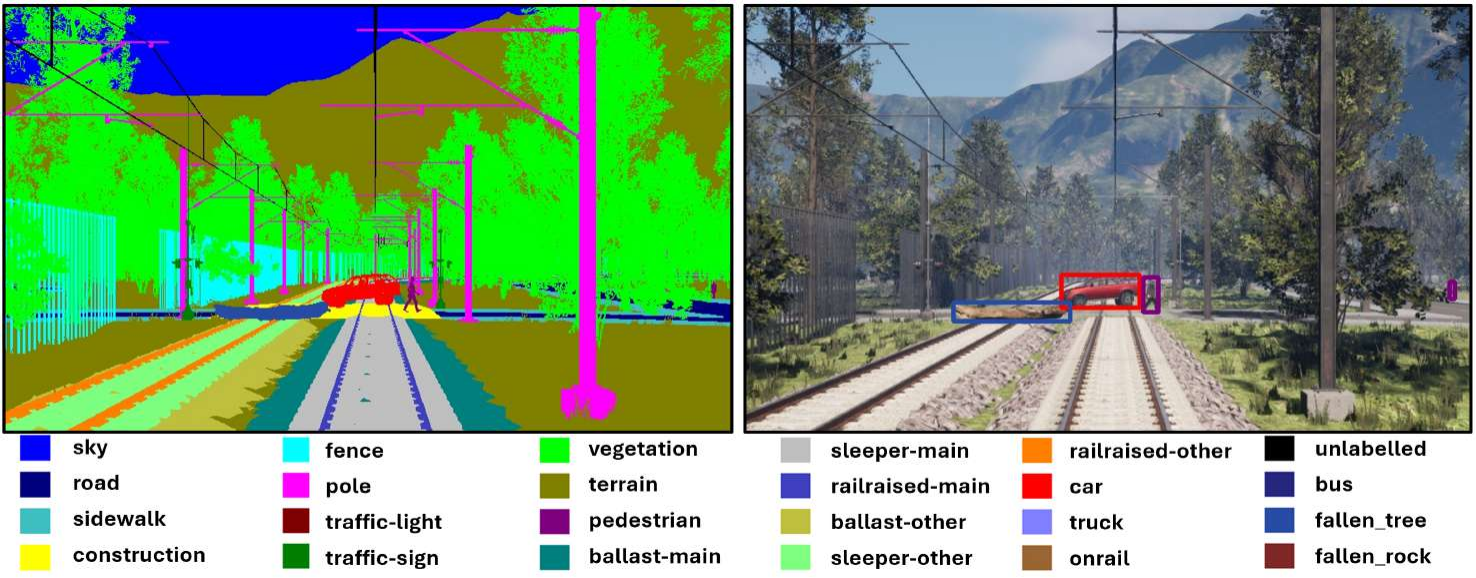}
\caption{Samples from \name~showing a semantically segmented image on the left, which includes the proper legend for the classes, and an RGB image with relevant 2D bounding boxes on the right; both come from the same scene as the point cloud in Fig.~\ref{fig:point_cloud} and use the same semantic color mapping.}
\label{fig:two_images}
\end{figure*}
To overcome these limitations and integrate well-performing perception methods in a railway setting, simulations and domain adaptation techniques introduce promising solutions. 
In the automotive field, simulators are widely used to generate annotated datasets, including rare and dangerous corner case scenarios~\cite{dosovitskiy2017carla, nvidia_drivesim}.
Moreover, domain adaptation methods have proven effective in transferring models trained on synthetic data to real-world applications~\cite{SSDA3D_2023}. 
These techniques can similarly be used to support the railway sector, both by adapting models trained on synthetic data to real railway environments and by transferring knowledge from automotive datasets to railway-specific scenarios.

In contrast to the automotive and robotics sectors, where tools like CARLA~\cite{dosovitskiy2017carla}, AirSim~\cite{shah2018airsim}, or NVIDIA Drive Sim~\cite{nvidia_drivesim} are commonly used to generate synthetic data, the railway domain has far fewer options ~\cite{d2023trainsim, de2023scenario, toprak2020conditional}.
Among the few recent efforts, Iglesias et al.~\cite{iglesias2024enhancing} present a CARLA-based approach for generating synthetic railway data, however, since the work is still under review, it cannot be used to test the proposed methods.

Even though domain adaptation methods can help bridge the gap between domains (both in sim-to-real and automotive-to-railway scenarios) the following challenges remain:
\begin{enumerate}
    \item The sim-to-real domain shift often limits the direct transferability of models trained on synthetic data to real-world environments, since simulations usually introduce "perfect" data, when in the real-world data noise is commonplace.
    \item While automotive datasets typically focus on dense, urban settings, railway environments are often open-field, sparse, and operate on different spatial and semantic scales, further complicating adaptation.
\end{enumerate}
Given the synthetic data and domain adaptation challenges, the contribution of this paper is threefold:
\begin{enumerate}
    \item We introduce \name, an extension of the synthetic dataset \originalname~\cite{d2025syndra}, which includes camera, depth and LiDAR data, along with multiple annotations, to support the evaluation of vision-based algorithms in railway environments. 
    To the best of our knowledge, \name~is the first publicly available synthetic dataset that supports both 2D and 3D object detection and semantic segmentation tasks in this domain.
    Figure~\ref{fig:two_images} presents an example RGB image alongside its corresponding semantic segmentation, both captured from a virtual environment generated within the \name~framework.
    \item We apply and adapt a state-of-the-art semi-supervised domain adaptation approach for 3D point clouds (SSDA3D) to evaluate both the transferability of models trained on \name~to real railway data and the impact of incorporating automotive data (Waymo) for cross-domain adaptation.
    \item We provide an analytical evaluation of how synthetic railway data, real-world automotive data, and their combination influence domain adaptation performance when transferring to a real-world railway dataset (OSDaR23). Moreover the original SSDA3D focused only on the car class, but we report results also for pedestrian detection since that is the most vulnerable traffic actor.
\end{enumerate}

The rest of the paper is organized as follows: Section~\ref{sec:relatedwork} presents the related work in this field;
Section~\ref{sec:syndra} introduces the \name~dataset;
Section~\ref{sec:3dssda} describes our optimized SSDA3D domain adaptation method;
Section~\ref{sec:exp} presents the experiments carried out to
evaluate the proposed domain adaptation framework across various training setups; and Section~\ref{sec:conc} states the conclusions and future work. 

\section{Related Works}\label{sec:relatedwork}

This section reviews existing open datasets for automotive and railway domain, focusing on those containing labels in the form of 3D bounding boxes. It then introduces domain adaptation, in particular describes a specific method like SSDA3D which enables the transfer of 3D object detection models across different domains and environments.

\subsection{Vision-based Datasets}

The automotive field has seen the release of several public large-scale datasets that support vision based tasks such as 2D/3D object detection task using cameras and LiDAR sensors.
Among them, KITTI~\cite{geiger2012we}, SemanticKITTI~\cite{behley2019semantickitti}, nuScenes~\cite{caesar2020nuscenes}, and the Waymo Open Dataset~\cite{sun2020scalability} stand out for their completeness, dense annotations, and rich sensor configurations.
In particular, Waymo provides high-resolution camera and LiDAR data with accurate 3D bounding boxes annotations for vehicles, pedestrian, and cyclists.
Due to its quality, scale, and widespread use as a benchmark for 3D perception tasks and domain adaptation techniques, it is used as real-world source domain in the comparison with the proposed synthetic data.

By contrast, publicly available datasets are limited in the railway domain, especially those offering annotated 3D point clouds or 3D bounding boxes.
For instace, datasets such as RailGoerl24~\cite{tagiew2025railgoerl24} RailSem19~\cite{zendel2019railsem19}, RailDet~\cite{cao2022effective}, and RailSet~\cite{zouaoui2022railset} provide valuable resources for 2D vision tasks, but lack the 3D annotations required for object detection algorithms.

Given the difficulty and cost of collecting real-world railway data, synthetic datasets have emerged also in this field as a promising alternative.
For example, de Gordoa et al.~\cite{de2023scenario} extended the CARLA simulator~\cite{dosovitskiy2017carla} to generate synthetic railway images under diverse operational conditions, while D'Amico et al.~\cite{d2023trainsim} introduced TrainSim, a simulation framework based on Unreal Engine 4 capable of producing images and point clouds for different vision-based tasks, but not providing 2D or 3D bounding box annotations.
Other simulation-driven datasets, such as RAWPED~\cite{toprak2020conditional}, RailEnV-PASMVS~\cite{broekman2021railenv}, and SARD~\cite{neri20223d}, target specific application like 2D object detection, geometry reconstruction, or railway signal classification, but do not provide annotated bounding boxes for point cloud-based object detection.
Finally, to the best of our knowledge, OSDaR23~\cite{tagiew2023osdar23} is the most complete public 3D object detection dataset for real-world railway applications including multi-sensor dataset. 
It includes 45 sequences comprising synchronized data from nine RGB cameras, one radar, and six LiDAR sensors, with annotations such as 3D bounding boxes and rail polylines, although a significant number of sequences are recorded while the train is stationary, reducing temporal diversity.

\subsection{Domain Adaptation for 3D Object Detection} 

Domain adaptation in 3D object detection refers to techniques that enable models trained on one data distribution (source domain) to generalize effectively on data having a different but related distribution (target domain).

Among semi-supervised domain adaptation methods for 3D object detection, SSDA3D~\cite{SSDA3D_2023} stands out by effectively leveraging both labeled and unlabeled target data alongside fully labeled source data. It addresses the challenges posed by large domain shifts between datasets, such as sensor differences and scene variability, through a novel two-stage training strategy.

The first stage, Inter-domain Point-CutMix, performs domain alignment by cutting and pasting spatially coherent regions between source and target point clouds. This encourages learning of domain-invariant representations by exposing the model to hybrid samples that contain mixed domain characteristics, thus making the detector more robust and reducing distribution discrepancies in 3D feature space.

The second stage, Intra-domain Point-MixUp, regularizes training on unlabeled target data by mixing pairs of actual target samples and pseudo-labels. This interpolation promotes consistent model predictions on unseen samples and mitigates the impact of pseudo-label noise. As a result, SSDA3D surpasses the fully supervised oracle - i.e. the model trained solely on the whole target dataset - using only a small fraction of target labels, as proved in Waymo-to-nuScenes~\cite{SSDA3D_2023} adaptation.

In summary, we chose OSDaR23 due to its comprehensive coverage and its status as one of the few publicly available railway datasets. We selected and optimized SSDA3D because of its demonstrated effectiveness in domain adaptation scenarios. Furthermore, we introduced \name~to address the lack of versatile, railway-focused datasets and to advance research in this underexplored domain.

\section{\name}\label{sec:syndra}

\originalname~\cite{d2025syndra} is a synthetic dataset for railway applications based on Unreal Engine 5, designed to support semantic image segmentation under different light and weather conditions. The proposed
\name~dataset extends \originalname~by introducing annotations specifically designed for 2D and 3D object detection in railway environments. The rest of this section describes the main features of the dataset.

\subsection{Virtual Environments}
\name~comprises seven distinct level crossing scenarios. 
For each scenario, we collected multiple image sequences, each focusing on a specific relevant object type (e.g., vehicle, pedestrian, or natural obstacle) crossing or occupying the rails for detection, while other static or dynamic elements are present in the scene.
Additionally, we created a bonus scenario in a large railway station environment, where only pedestrians cross the tracks.
\begin{figure}[htbp]
\centering
\begin{subfigure}[b]{0.24\textwidth}
    \centering
    \includegraphics[width=\textwidth]{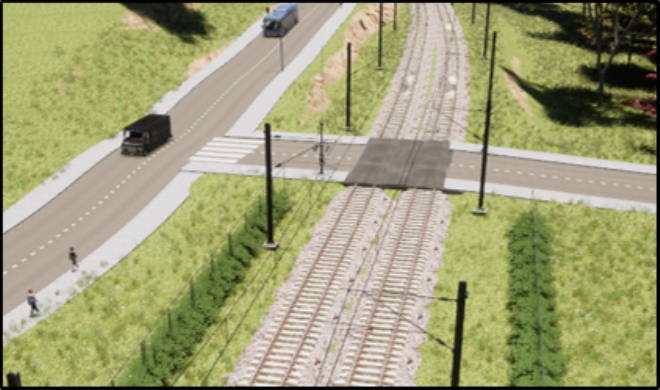}
\end{subfigure}
\hfill
\begin{subfigure}[b]{0.24\textwidth}
    \centering
    \includegraphics[width=\textwidth]{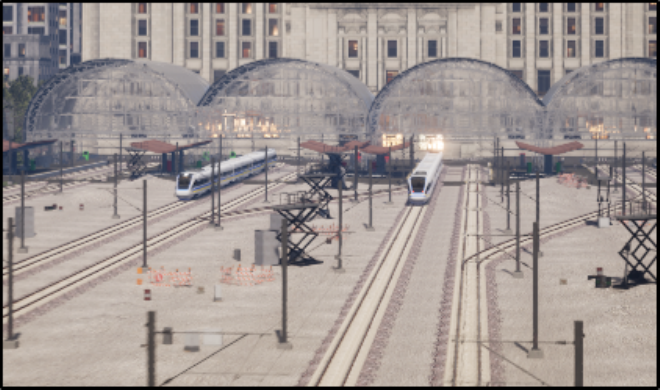}
\end{subfigure}
\caption{Example scenarios from \name. The left image shows an open-field scene near a level crossing with moving objects such as vehicles and pedestrians. The right image presents a sample from the bonus scenario, depicting a detailed railway station environment.}
\label{fig:scenarios}
\end{figure}
Figure~\ref{fig:scenarios} depicts one of the seven virtual environments alongside the bonus railway station scenario included in \name.

In each sequence, the train starts approximately 350 meters from the level crossing. 
This setup enables the simulation of a range of realistic hazardous situations, with one primary obstacle type highlighted per sequence, allowing for focused and selective evaluation of the 3D perception system under test.
The obstacles include:
\begin{itemize} 
    \item \textit{Vehicles}: three sequences featuring a truck, a car, and a bus crossing the railway at different times and distances from the approaching train.
    \item \textit{Pedestrians}: one sequence with two pedestrians crossing the tracks in opposite directions (left-to-right and right-to-left), and another sequence with two pedestrians walking parallel to the tracks, one moving in the same direction as the train, the other in the opposite.
    \item \textit{Natural obstacles}: three sequences containing a fallen tree with leaves, bare branches, and rocks obstructing either the main or adjacent track.
\end{itemize}
This per-sequence obstacle configuration was designed to replicate realistic, safety-critical scenarios commonly encountered in rail transport. 
It ensures a diverse and controlled dataset structure, facilitating systematic and reproducible evaluation of perception models under varied but representative conditions.

\subsection{Sensor Configuration}

\name~provides high-resolution RGB images and synthetic LiDAR point clouds, along with detailed annotations for both semantic segmentation and bounding boxes in both the image and point cloud domains.

To support diverse research needs, each scene is rendered with two different camera field-of-view (FoV) settings, 30$^{\circ}$ and 90$^{\circ}$ for narrow and wide perspectives, while maintaining the same image resolution of  $2464\times1600$. 
Similarly, both TELE-15~\footnote{https://www.livoxtech.com/tele-15} and 64-beam Velodyne HDL-64E~\footnote{ https://www.mapix.com/wp-content/uploads/2018/07/63-9194\_Rev-J\_HDL-64E\_S3\_Spec-Sheet-Web.pdf} point clouds are provided to allow evaluation across different LiDAR configurations.

\subsection{2D/3D Box Annotation Pipeline}

The annotation pipeline for \name~extends the original \originalname~dataset by providing accurate 2D and 3D bounding boxes for key object classes in railway scenarios. 
While the generation of semantic segmentation labels and the name convention of sensory data has been fully described in the original \originalname~paper, \name~follows the same labeling policy for point cloud segmentation, ensuring consistency in class definitions and annotation format. 
The class taxonomy is further refined for object detection tasks: generic vehicle classes are split into \textit{car}, \textit{bus}, and \textit{truck}, and new obstacle categories such as \textit{fallen\_tree} and \textit{fallen\_rock} are introduced to represent natural hazards commonly found in rail environments.

\subsection{3D bounding boxes}

3D bounding boxes are directly obtained from the UE5’s internal functions and are assigned to all objects belonging to detection-relevant classes (e.g., cars, trucks, pedestrians), independently of the number of LiDAR returns they generate. 
Although this choice may introduce cases where annotated objects are sparse or not visible at all in the point cloud, it enables a realistic evaluation of the selected LiDAR configuration in terms of both coverage and detection reliability.

The \name~dataset is publicly available for research purposes. 
It can be accessed and downloaded from the official project website\footnote{\dataseturl}, where further information on 2D/3D annotations and other sensor configurations can be found.
In addition to the original \originalname~dataset, the website includes all \name~acquisition sequences, RGB and depth images, LiDAR point clouds, and corresponding 2D/3D annotations in JSON format. 
A comprehensive documentation is also provided to assist the user with dataset usage, along with scripts for parsing and visualization purposes. 
Researchers are encouraged to cite this work when using \name~in their experiments.

\section{Optimizing SSDA3D for Railway Scenes}\label{sec:3dssda}

As discussed in the previous sections, this work is based on SSDA3D. 
The core of this domain adaptation pipeline lies in the \textbf{CutMix} and the \textbf{PointMixUp} modules, as well as its ability to outperform oracle models.
Note that SSDA3D was originally designed to handle data distribution shifts between real-world automotive datasets (e.g., Waymo to nuScenes), while here we address two other types of shifts: (a) road-to-rail, meaning real automotive to real railway data, and (b) sim-to-real, meaning synthetic to real railway data.
To reduce the impact of such shifts during the training of a SSDA3D-based pipeline, several adjustments were made.

First, we focus on long-range detection, as it’s critical for long train braking distances.
Therefore, we use the TELE-15 point clouds from \name~and filter OSDaR23 to retain only the points of the TELE-15 sensor in the middle, removing data from other LiDARs to ensure alignment.

Second, in the original SSDA3D the \textbf{CutMix} operation is done by taking some region of a point cloud from the target dataset and pasting it into a point cloud from the source dataset. 
Perhaps the cut-paste order between the source and target datasets does not matter much, as both are from the common automotive sector, meaning the background points resemble a road-like environment regardless of which dataset takes the role of source or target.
However, in our work the opposite was done: a region from the source was pasted into the target, since Waymo as the source domain lacks the same structural and spatial layout observed in the OSDaR23 railway setting.
In this way, after \textbf{CutMix} the resulting point clouds still retains the characteristic railway surroundings shape as the main desired region of interest.

Third, we are dealing with very different FOVs between the source Waymo ($360^\circ$) and target OSDaR23 ($15^\circ$). 
The narrow field of view of the TELE-15 sensor leads to point clouds with a sparse and elongated frustum, resulting in a detection range that has a rectangular shape in BEV with large empty space.
In the original \textbf{CutMix}, the cropped source points are pasted into the target at the exact same location with equal coordinates, which may lead to unrealistic placements due to differing occupied areas between source and target point clouds. Our modified approach mitigates this by computing a horizontal translation vector from the source crop center to the nearest target point. Before pasting, the source points are shifted by this vector, aligning the source crop with the target’s corresponding location. This process ensures the source cut region is embedded realistically within the target point cloud.

\begin{figure}[htbp]
\centering
\begin{subfigure}[b]{0.43\textwidth}
    \centering
    \includegraphics[width=\textwidth]{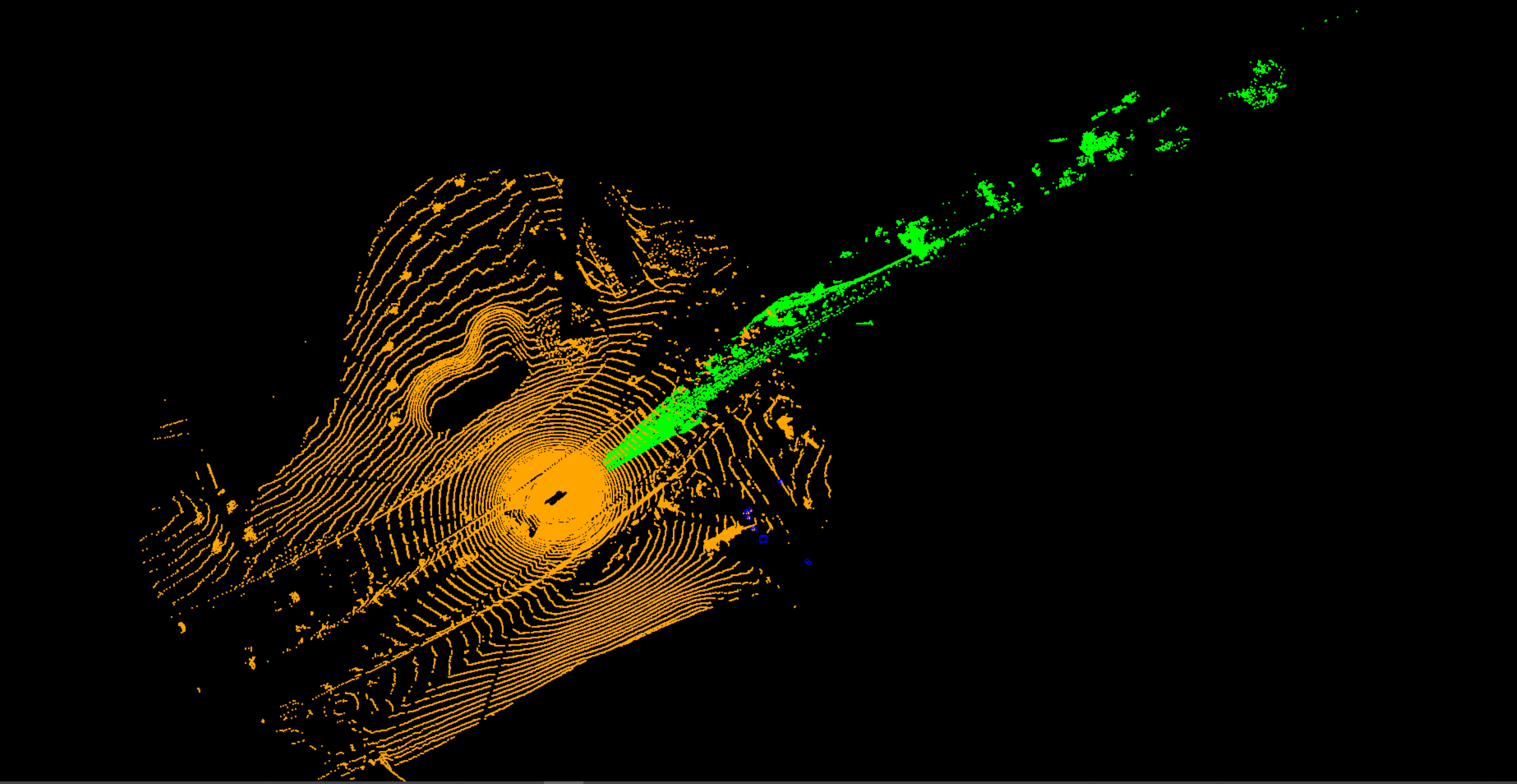}
    \caption{Waymo and OSDaR23 point clouds overlap}
    \label{fig:WO_pcds}
\end{subfigure}

\begin{subfigure}[b]{0.43\textwidth}
    \centering
    \includegraphics[width=\textwidth]{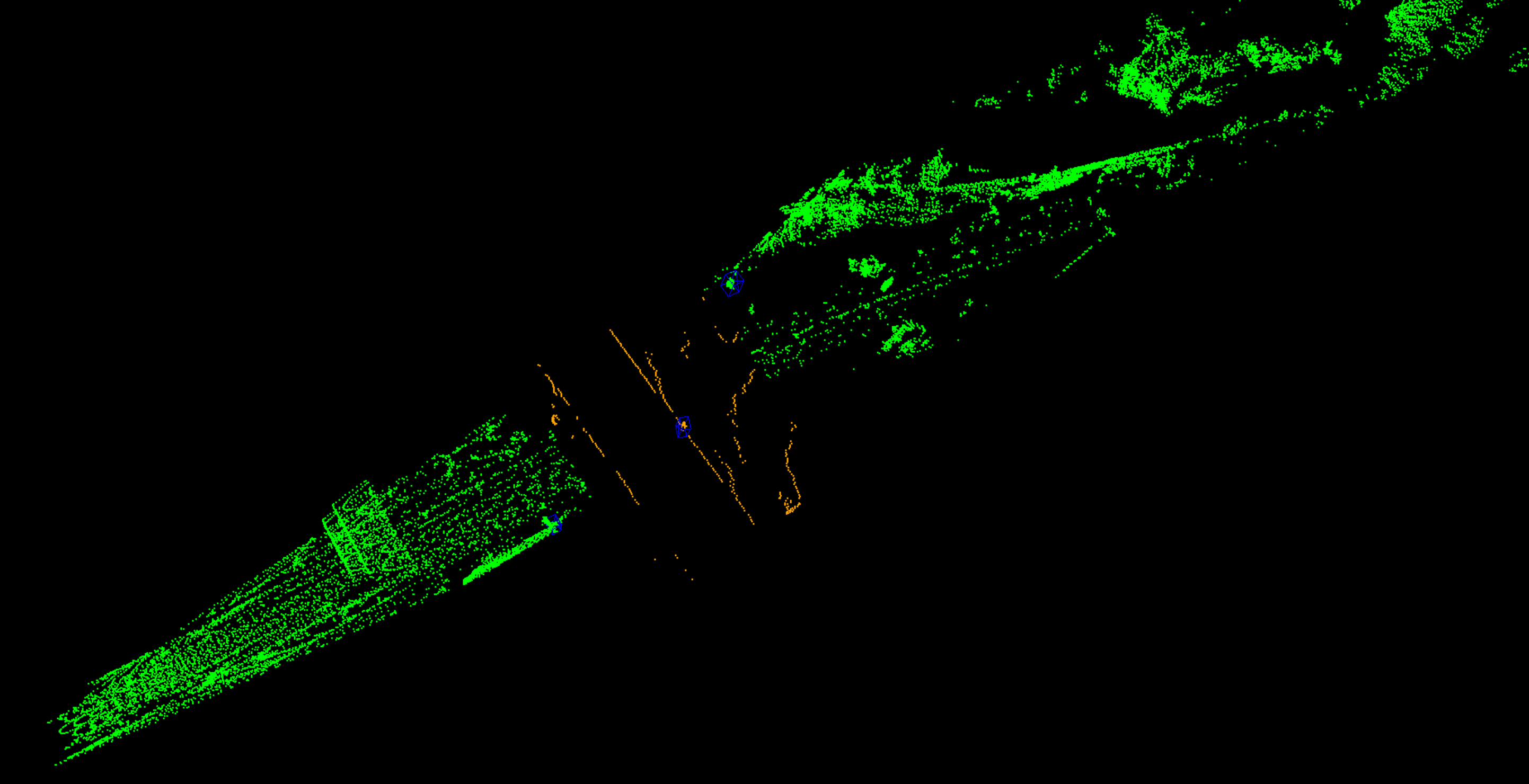}
    \caption{\textbf{CutMix} hybrid point cloud}
    \label{fig:CutMix_result}
\end{subfigure}

\caption{Waymo (orange) and OSDaR23 (green) point clouds before (a) and after (b) the \textbf{CutMix} operation. In (b) only a small cut region from the source Waymo frame was pasted into the target OSDaR23 frame. Boxes in blue are pedestrian labels. Best viewed zoomed in.}
\label{fig:CutMix_mosaic}
\end{figure}

Fourth, the original \textbf{CutMix} strategy is pure random and does not guarantee that the cut region contains at least one object label. 
This is acceptable for large datasets, but problematic for small ones like OSDaR23, where underrepresented classes (e.g., cars) make every example valuable.
For this reason, a check step was introduced in the proposed \textbf{CutMix} pipeline: before cutting any region from the source cloud, we verify whether the region contains at least one ground-truth box. Figure \ref{fig:CutMix_mosaic} illustrates the effect of our custom \textbf{CutMix} operation.

Fifth, although SSDA3D was originally designed to operate with a single source dataset, this work investigated the use of multiple sources, combining large-scale real-world automotive data with synthetic railway data to improve generalization when training on the OSDaR23 target domain, which is a rather small dataset that contains a limited number of annotated point clouds with very low variability.
While training SSDA3D on a single large source dataset such as Waymo already outperforms a detector like CenterPoint~\cite{Centerpoints} trained on the target domain alone, combining datasets offers additional benefits. 
The Waymo dataset contains large amounts of diverse, annotated labels for objects, while \name~brings structural similarity to the target railway domain.
A naïve implementation would apply \textbf{CutMix} with a 0.5 probability of selecting a region from either source dataset. 
However, this approach can be problematic when one dataset is significantly smaller than the other. 
Repeated sampling from the smaller dataset may lead to redundancy and under-utilization of the larger dataset.
To address this problem, a size-aware \textbf{CutMix} sampling strategy is proposed, where the probability of selecting a dataset during \textbf{CutMix} is proportional to its size.
This allows the model to benefit from the volume of the larger real dataset while still incorporating samples from the smaller domain-specific synthetic source.
\section{Experiments}\label{sec:exp}
This section presents the experimental setup, including dataset configurations, evaluation metrics, implementation details, and performance comparisons.

\subsection{Datasets}
The target domain is OSDaR23, a real-world public railway dataset, while the synthetic source is \name, which offers critical railway scenarios. We also use the Waymo Open Dataset (WOD), an automotive dataset.
The motivation for adding a cross-domain dataset is its volume and diversity, helping mitigate the scarcity of real labeled railway data

Since these datasets have different object categories, only the common object classes are considered: \textit{Person} / \textit{Pedestrian} and \textit{Car} / \textit{Vehicle}, selected for availability and relevance.
Figure~\ref{tab:dataset_stats} shows statistics about the employed datasets.
\name~was split into three subsets: $70\%$ for training, $20\%$ for validation, and $10\%$ for testing, whereas OSDaR23 and WOD were split using their predefined divisions.
Frame split for \name~was done in 10-frame batches: every 6th frame to test, every 3rd and 9th to validation, the rest to training.

During training, we use $100\%$ of each training set. 
\textbf{CutMix} is applied with a $30\%$ probability, and MixUp with $50\%$. 
Additionally, \name~provides annotations even for objects that generate a single LiDAR point return. 
However, to ensure consistency with real-world datasets and focus on reliably detectable instances, we retain only objects with at least five points.

\begin{table}[!hbtp]
\centering
\small
\begin{tabular}{|c|ccc|cc|}
\hline
\multirow{2}{*}{\textbf{Dataset}} &
\multicolumn{3}{c|}{\textbf{Frames}} &
\multicolumn{2}{c|}{\textbf{Label Count}} \\
 & Train & Val & Test & Car & Pedestrian  \\ \hline
OSDaR23     & 778 & 189 & 160 & 12.669 & 73.421  \\ \hline
\name & 4.838 & 1.367 & 672 & 14.359 & 12.107  \\ \hline
Waymo       & 158k & 40k & 30k & 6.024k & 2.772k  \\ \hline
\end{tabular}
\caption{Overview of dataset statistics: frame counts for training, validation, and test splits; plus Car / Pedestrian label counts across datasets.}
\label{tab:dataset_stats}
\end{table}

\begin{table*}[!hbtp]
\centering
\setlength{\tabcolsep}{1pt}
\resizebox{2\columnwidth}{!}{%
\begin{tabular}{|c|c|cccccc|cccccc|}
\hline
 \multirow{3}{*}{\textbf{Method}} & 
  \multirow{3}{*}{\textbf{Stage}} &
  \multicolumn{6}{c|}{\textbf{Car}} &
  \multicolumn{6}{c|}{\textbf{Person}} \\ \cline{3-14} 
 &  &
  \multicolumn{3}{c|}{\textbf{0.7}} &
  \multicolumn{3}{c|}{\textbf{0.5}} &
  \multicolumn{3}{c|}{\textbf{0.5}} &
  \multicolumn{3}{c|}{\textbf{0.25}} \\ \cline{3-8} \cline{9-14}
 & 
   &
  \textbf{AP BEV} &
  \textbf{AP 3D} &
  \textbf{Closed Gap} &
  \textbf{AP BEV} &
  \textbf{AP 3D} &
  \textbf{Closed Gap} &
  \textbf{AP BEV} &
  \textbf{AP 3D} &
  \textbf{ClosedGap} &
  \textbf{AP BEV} &
  \textbf{AP 3D} &
  \textbf{Closed Gap} \\ \hline
\textbf{S-Only} & N/A & 0.0 & 0.0 & 0 / 0 & 0.0 & 0.0 & 0 / 0  & 0.0 & 0.0 & 0.0 / 0.0 & 0.73 & 0.64 & 0 / 0 \\ 

\textbf{W-Only} & N/A & 0.0 & 0.0 & 0 / 0 & 0.0 & 0.0 & 0 / 0 & 20.47 & 0.0 & 0 / 0 & 35.38 & 0.002 & 0 / 0 \\ 

\textbf{Oracle} & N/A & 85.77 & 20.73 & 100 / 100 & 85.77 & 48.26 & 100 / 100 & 42.13 & 20.21 & 100 / 100 & 53.41 & 51.60 & 100 / 100 \\ 
\hline

\textbf{S$\rightarrow$O} & CutMix & 83.75 & 6.61 & 97.64 / 31.89 & 83.75 & 57.27 & 97.64 / 118.7 & 41.07 & 24.89 & 97.48 / 123.16 & 54.86 & 52.80 & 102.75 / 102.35 \\ 
 & MixUp & 85.49 & 1.72 & 99.67 / 7.72 & 85.49 & 66.26 & 99.67 / 137.3 & 40.53 & 29.58 & 96.2 / 146.36 & 54.66 & 54.66 & 102.37 / 102.37 \\ 
\hline

\textbf{W$\rightarrow$O} & CutMix & 84.72 & 8.21 & 98.78 / 39.6 & 84.72 & \textbf{70.73} & 98.78 / \textbf{146.6} & \textbf{48.77} & 32.45 & \textbf{130.66} / 160.56 & 59.34 & 59.18 & 132.89 / 114.69 \\ 
 & MixUp & \textbf{88.21} & 1.6 & \textbf{102.84} / 7.71 & \textbf{88.21} & 69.76 & \textbf{102.84} / 144.55 & 46.81 & 32.90 & 121.60 / 162.79 & 63.85 & 63.71 & \textbf{157.90} / 123.47 \\ 
\hline

\textbf{SW$\rightarrow$O} & CutMix & 84.29 & \textbf{10.31} & 98.27 / \textbf{49.73} & 84.29 & 59.86 & 98.27 / 124.03 & 48.72 & 33.23 & 115.64 / 164.42 & 61.46 & 61.07 &  115.07 / 118.35 \\ 
 & MixUp & 85.10 & 0.68 & 99.22 / 3.3 & 85.10 & 59.56 & 99.22 / 123.41 & 45.08 & \textbf{35.67} &  107.00 / \textbf{176.50} & \textbf{71.08} & \textbf{70.35} & 133.08 / \textbf{136.34} \\ 
\hline

\end{tabular}%
}
\caption{Domain adaptation performance for all the experiments using \name~/ Waymo / a combination of both as source and OSDaR23 as the target. We report AP BEV, AP 3D and their corresponding Closed Gap on the \textit{Car} and \textit{Person} classes for all methods. All reported results are expressed as percentages ($\%$). Bold values indicate the highest results in each column.}
\label{tab:exp_results}
\end{table*}
\subsection{Evaluation Metrics}
Model performance was evaluated using 3D Average Precision (AP) and Average Precision in Bird’s-Eye-View  (AP BEV), which measure the accuracy of predicted 3D bounding boxes relative to ground-truth annotations.
Metrics were computed at various IoU thresholds: 0.7 and 0.5 for cars, and 0.5 and 0.25 for persons. 
For each threshold, the AP is reported using the AP40 metric defined in KITTI.

Additionally, following~\cite{yang2021st3d}, the Closed Gap metric was used to quantify how much of the performance gap between the source-only and oracle models is closed through adaptation:
\begin{equation}
    \mbox{\textbf{Closed Gap}} = \frac{AP_{model}-AP_{source-only}}{AP_{oracle}-AP_{source-only}} \times 100\%
\end{equation}

\subsection{Baseline and Comparison}
The baseline model is CenterPoint~\cite{Centerpoints}, as in the original SSDA3D.
The experiments include the following training setups:
\begin{itemize}
    \item \textbf{S-only, W-only}: CenterPoint trained only on source data, i.e. \name~or Waymo, and evaluated on OSDaR23, to quantify the domain gap.
    \item \textbf{Oracle}: CenterPoint trained and evaluated on OSDaR23 with full supervision, upper-bound performance.
    \item {\textbf{S$\rightarrow$O}}: CenterPoint is trained using our modified SSDA3D and only uses a synthetic-to-real domain adaptation, with \name~as the source and OSDaR23 as the target.
    \item {\textbf{W$\rightarrow$O}}: Similar to the previous setting, but using a cross-industry domain adaptation with Waymo as the source instead.
    \item {\textbf{SW$\rightarrow$O}}: CenterPoint is trained with both domain adaptation techniques where \name~and WOD are the sources, and OSDaR23 is the target.
\end{itemize}

\subsection{Implementation}
All models were implemented using the OpenPCDet framework \cite{openpcdet2020}. 
Source-only and oracle models were trained for 20 epochs. 
For the adaptation setups, 20 epochs were used for stage one, followed by 20 additional epochs for stage two. 
The detection range was set to [0.0, -54.0, -3.0, 216.0, 54.0, 6.8] meters.
Additionally, Waymo and OSDaR23 provide an intensity channel as the fourth feature in each point, whereas, \name~does not currently include an intensity channel.
To handle this inconsistency, when using only WOD and/or OSDaR23, we used all four point features (XYZ + intensity); when using \name~alone or together with OSDaR23, we only used the XYZ coordinates; for SW$\rightarrow$O experiment, we used four channels, assigning a constant intensity value of 1 to \name~points to increase robustness.
All intensity values were normalized. Moreover, since Waymo and OSDaR23 provide point clouds already aligned to the ground in the vehicle frame, while \name~provides points in the sensor frame with the origin located a few meters above, we adjusted the point clouds in \name~to align with the ground level of the other two datasets.

\subsection{Results}
\subsubsection{Quantitative}
The experimental results in Table \ref{tab:exp_results} reveal compelling insights about cross-domain knowledge transfer in railway object detection. 
Surprisingly, adaptation from Waymo (automotive domain) to OSDaR23 (railway domain) outperforms \name~(synthetic railway) adaptation across most metrics, achieving a 146.6\% Closed Gap for cars in AP 3D at 0.5 IoU. 
This suggests that real-world automotive data contains transferable features more beneficial for railway detection than synthetic domain-specific data, potentially due to Waymo's rich diversity in real-world lighting conditions, textures, and occlusion patterns. 
The superiority of Waymo-based adaptation persists even for pedestrian detection, where it achieves 160.56\% Closed Gap compared to \name~ 123.16\% in AP 3D at 0.5 IoU, indicating that human appearance patterns in automotive contexts generalize better to railway environments than synthetically generated pedestrians.  
To assess domain differences, we analyzed the height and range statistics of both full point clouds and pedestrian-specific points. For all points, SynDRA-BBox exhibits higher mean height ($7.03 m \pm 7.52$) and range ($136.89 m \pm 118.25$) compared to OSDaR23 ($3.58 m \pm 3.61$, $68.36 m \pm 60.77$), indicating that SynDRA-BBox scenes are typically more open, while OSDaR23 is collected in more confined environments such as stations or vegetated areas.
For pedestrian points, SynDRA-BBox again shows greater average range ($110.62 m \pm 76.92$) but lower average height ($0.67 m \pm 0.43$), whereas OSDaR23 has closer ($49.86 m \pm 28.10$) and slightly higher ($0.77 m \pm 0.46$) pedestrian returns. This reflects both the open-field nature of SynDRA scenes and limitations in pedestrian modeling.
The combination of open spatial layouts, noise-free sensor simulation, and less detailed pedestrian geometry in SynDRA-BBox likely reduces the realism of point cloud patterns, contributing to the reduced adaptation performance when compared to real-world datasets like OSDaR23 or Waymo.

By contrast, the combined \name+Waymo approach (\textbf{SW$\rightarrow$O}) demonstrates synergistic benefits, particularly for decent pedestrian detection at 0.25 IoU ($136.34\%$ Closed Gap in AP 3D). 
This hybrid strategy likely succeeds by merging \name~domain-specific railway layout patterns with Waymo robust real-world object features. 
Notably, \textbf{SW$\rightarrow$O} with MixUp augmentation surpasses Oracle performance in multiple categories (e.g., $71.08$ AP BEV vs Oracle's $53.41$ for Person-0.25), suggesting that strategic domain combination can overcome limitations of both synthetic data and single-source adaptation. 
These findings highlight the value of hybrid domain adaptation frameworks in railway applications, particularly when combining real-world automotive data with synthetic domain-specific content to address the data scarcity challenges inherent in railway environments.

The results in Table~\ref{tab:ablation} highlight two important findings for domain adaptation in railway pedestrian detection. 
First, while domain adaptation using only Waymo (Tuned W$\rightarrow$O) provides solid performance, incorporating \name~as an additional synthetic source (SW$\rightarrow$O) leads to further improvements, especially when the adaptation framework is properly tuned. 
Notably, the "Tuned SW$\rightarrow$O" approach outperforms all other methods, achieving the highest AP 3D scores at both 0.5 ($35.67$) and 0.25 ($70.35$) IoU thresholds after the MixUp stage. 
This demonstrates the value of leveraging synthetic railway data in combination with real-world data to better bridge the domain gap and enhance detection performance.

\begin{table}[!hbtp]
\centering
\small
\begin{tabular}{|c|c|c|c|}
\hline
 Method & Stage & \textbf{0.5 AP 3D} & \textbf{0.25 AP 3D} \\ \hline
\multirow{2}{*}{\textbf{Tuned W$\rightarrow$O}} 
 & CutMix & 32.45 & 59.18 \\ 
 & MixUp  & 32.89 & 63.71 \\ \hline
\multirow{2}{*}{\textbf{Naive SW$\rightarrow$O}} 
 & CutMix & 30.99 & 51.53 \\ 
 & MixUp  & \textit{29.83} & \textit{58.4} \\ \hline
\multirow{2}{*}{\textbf{Tuned SW$\rightarrow$O}} 
 & CutMix & 33.23 & 61.07 \\ 
 & MixUp  & \textbf{35.67} & \textbf{70.35} \\ \hline
\end{tabular}
\caption{Comparison of domain adaptation performance on the Person class using our tuned SSDA3D vs. naive SSDA3D without CutMix adjustments. All reported results are expressed as percentages ($\%$). Best results (bold) and worst results (italic) are highlighted.}
\label{tab:ablation}
\end{table}
Moreover, the comparison between "Naive SW$\rightarrow$O" and "Tuned SW$\rightarrow$O" underscores the necessity of adapting and customizing the SSDA3D framework for the specific characteristics of the railway scenario. 
The tuned version consistently surpasses the naive, out-of-the-box application of SSDA3D, confirming that careful method tuning and domain-specific adjustments are critical for maximizing the benefits of multi-source domain adaptation in challenging real-world tasks.

\subsubsection{Qualitative}

\begin{figure}[htbp]
\centering
\begin{subfigure}[b]{0.24\textwidth}
    \centering
    \includegraphics[width=\textwidth]{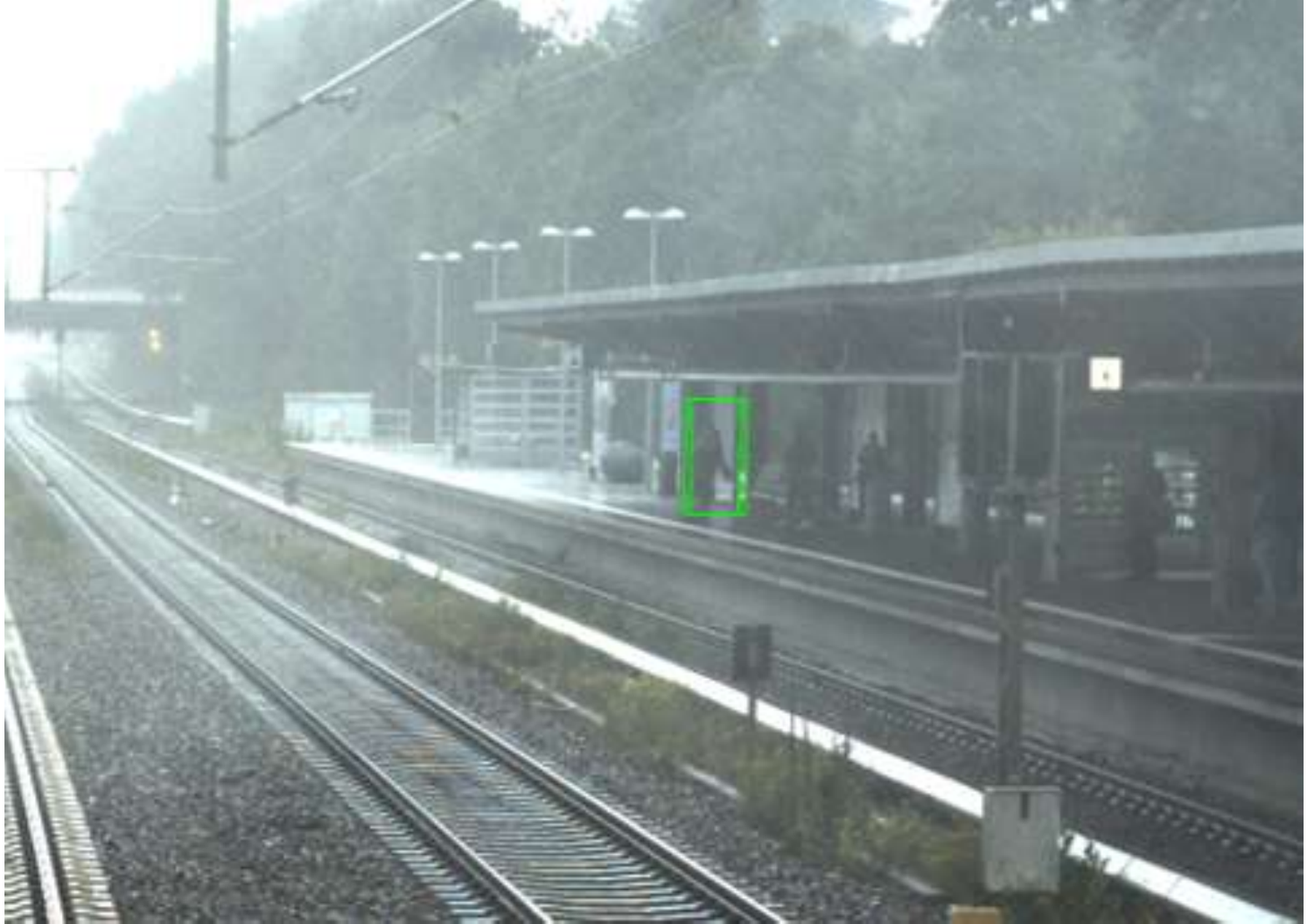}
    \caption{Oracle}
    \label{fig:oracle}
\end{subfigure}
\hfill
\begin{subfigure}[b]{0.24\textwidth}
    \centering
    \includegraphics[width=\textwidth]{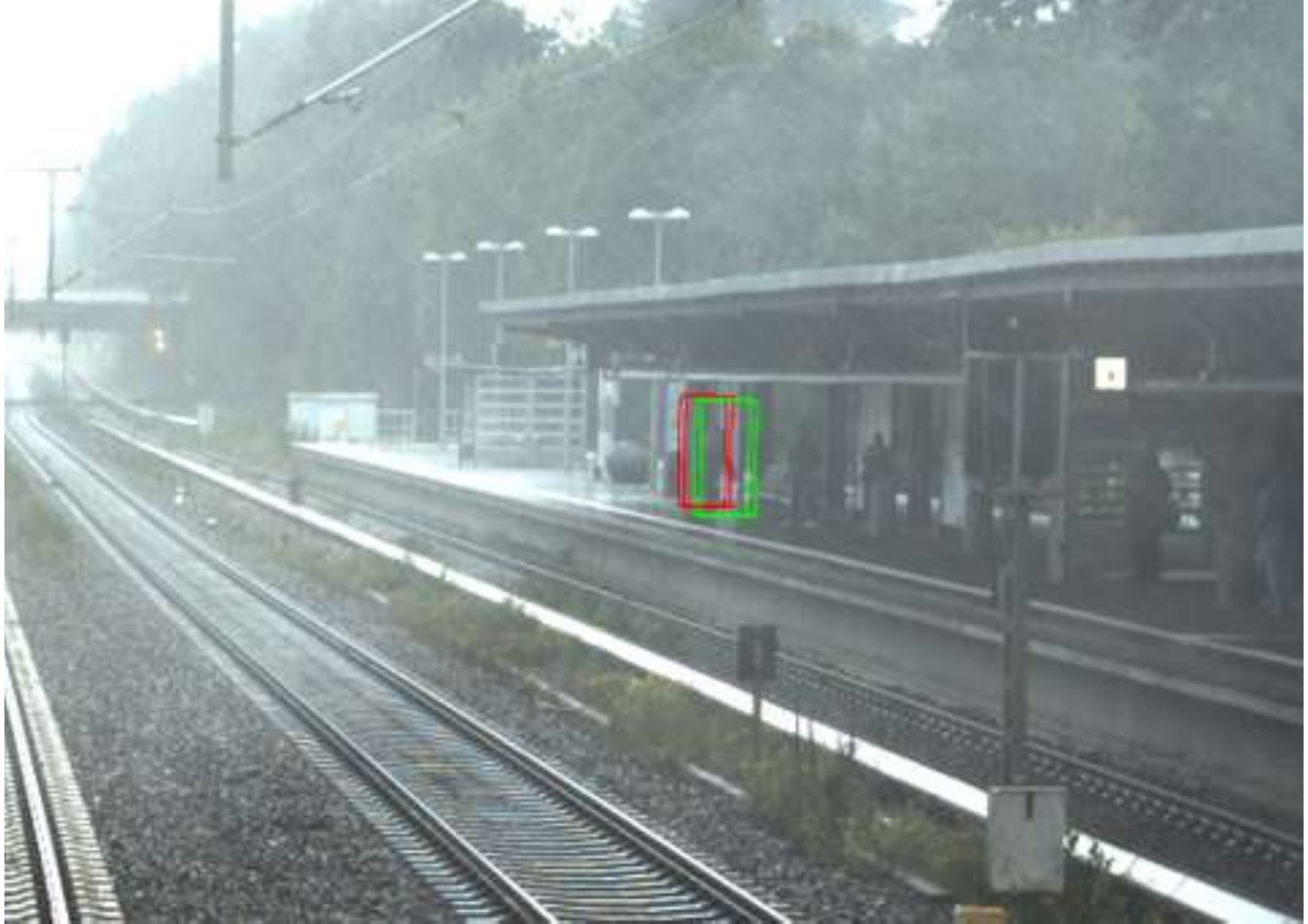}
    \caption{SW$\rightarrow$O}
    \label{fig:sinway1}
\end{subfigure}
\begin{subfigure}[b]{0.24\textwidth}
    \centering
    \includegraphics[width=\textwidth]{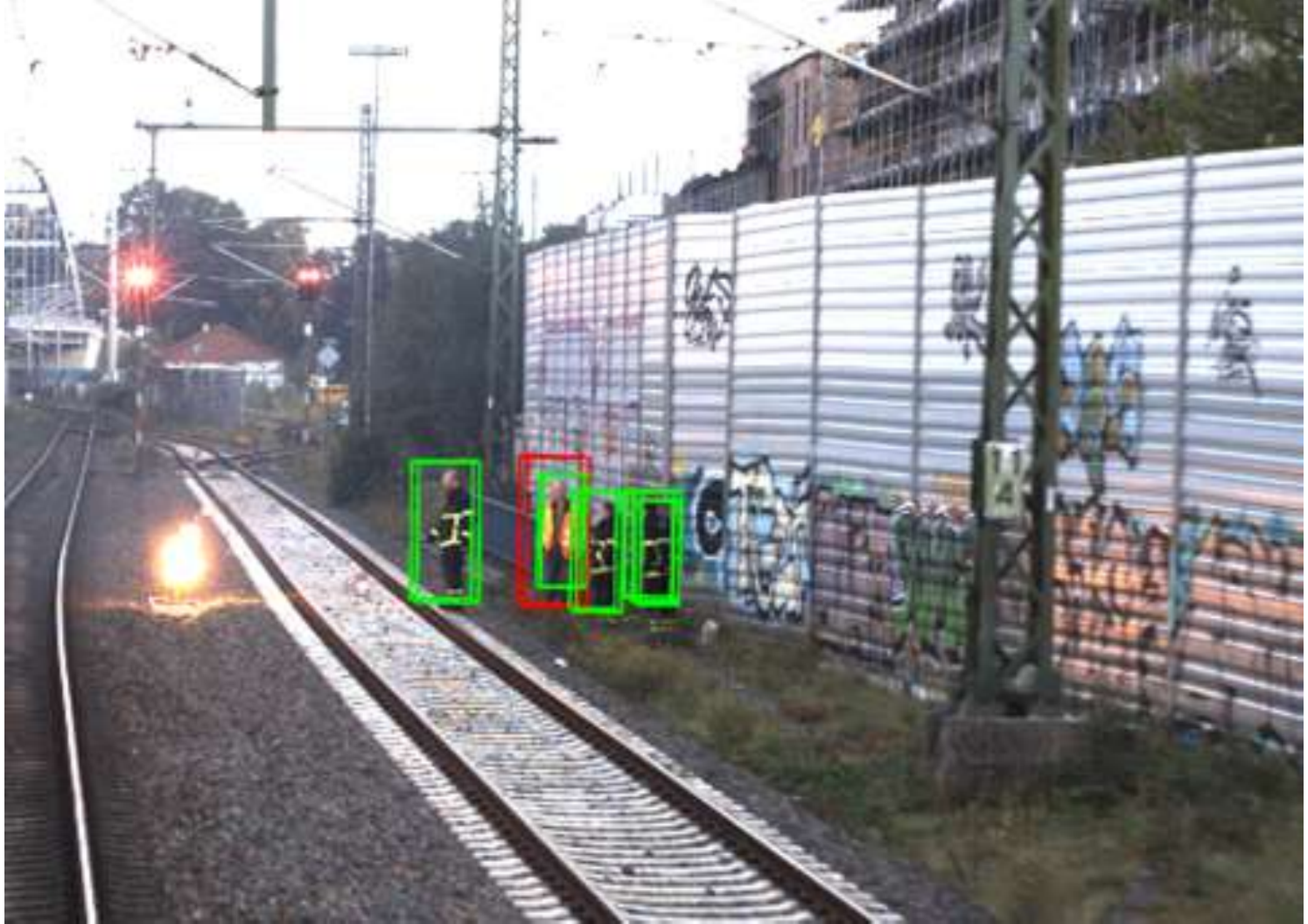}
    \caption{W$\rightarrow$O}
    \label{fig:waymo}
\end{subfigure}
\hfill
\begin{subfigure}[b]{0.24\textwidth}
    \centering
    \includegraphics[width=\textwidth]{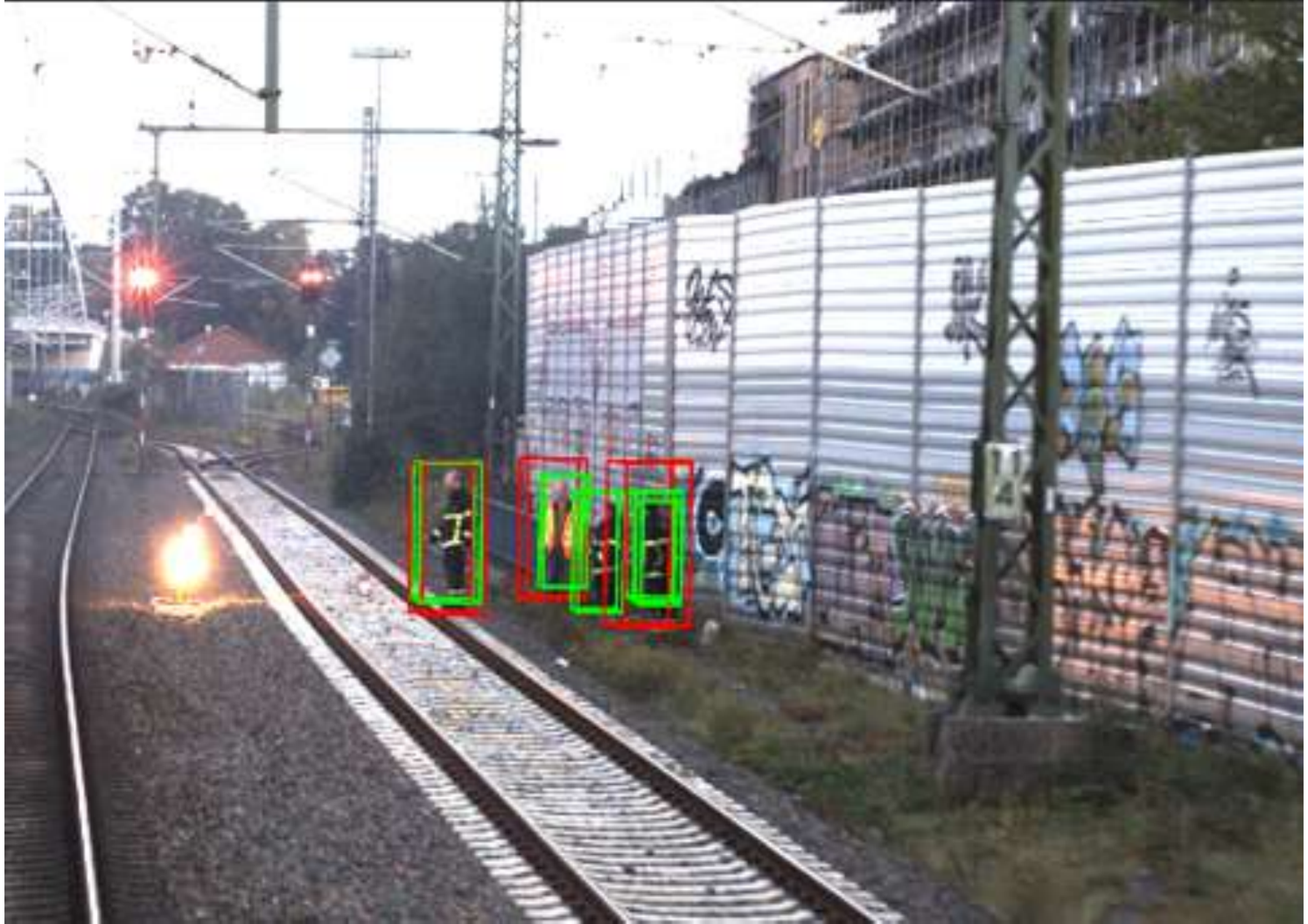}
    \caption{SW$\rightarrow$O}
    \label{fig:sinway2}
\end{subfigure}
\caption{Oracle vs. SSDA3D with both \name~and Waymo as sources and OSDaR23 as target. Ground truth boxes shown in green, predicted boxes in red.}
\label{fig:qual1}
\end{figure}

The qualitative results provide visual evidence of the benefits of advanced domain adaptation strategies for pedestrian detection in railway environments. 
In Figure~\ref{fig:oracle}, the Oracle model, despite full supervision, misses a pedestrian on the right platform, while the SSDA3D-adapted model (using both \name~and Waymo in Figure~\ref{fig:sinway1}) detects it even at long range.
This demonstrates that domain adaptation can enable the model to generalize better in difficult or underrepresented scenarios, sometimes even surpassing the supervised Oracle.
Furthermore, the added value of combining synthetic and real-world data for adaptation is illustrated in Figure~\ref{fig:waymo}.
When only Waymo is used as a source, the model detects just one person on the right track. 
By contrast, the inclusion of \name~alongside Waymo allows the model to detect three out of four pedestrians in the same scene, Figure~\ref{fig:sinway2}. 
These examples underline the importance of multi-source adaptation: integrating synthetic railway data with diverse real-world data leads to richer feature representations and improved detection robustness in complex or cluttered railway scenes. 

Therefore, the qualitative findings strongly support and emphasize that tailored domain adaptation strategies significantly enhance pedestrian detection performance in challenging real-world railway applications.
\section{Conclusion}\label{sec:conc}
A key challenge in advancing vision-based algorithms for railway environments is the lack of public datasets supporting tasks like 2D/3D object detection and semantic segmentation. To address this, we introduced \name, an extension of \originalname, offering multimodal data (camera, depth, LiDAR) with rich annotations. 
To our knowledge, \name~is the first publicly available synthetic dataset, accessible at \dataseturl, designed specifically for both detection and segmentation in railway scenarios, providing a foundation for future research in image and point cloud domains.

Building on this, we tuned and optimized SSDA3D for 3D object detection adaptation in rail settings, evaluating it across synthetic-to-real (\name~to OSDaR23) and cross-domain (Waymo to OSDaR23) setups.
Our results show that while synthetic data alone improves performance, cross-industry adaptation using Waymo outperforms synthetic-only setups, achieving up to $146.6\%$ Closed Gap for cars and $160.56\%$ for pedestrians, thanks to the diversity and richness of real-world features. Importantly, adding a dataset like \name~on top of Waymo further increases accuracy, particularly for challenging pedestrian cases (e.g., $176.50\%$ Closed Gap at 0.5 IoU), highlighting the added benefit of multi-source adaptation.
Qualitative analyses reinforce these findings: using both synthetic and automotive data, the model detects distant or underrepresented pedestrians that even the fully supervised Oracle misses, highlighting how tailored adaptation strategies can surpass baseline supervised models in complex railway scenes.

At last, we outline several future directions: developing adaptation strategies tailored to railway-specific structure and constraints beyond current methods like SSDA3D; extending SynDRA-BBox with diverse weather and lighting conditions (e.g., fog, rain, nighttime) to better reflect real operational environments; improving 3D pedestrian models and collision geometry, as Waymo's stronger performance suggests the importance of shape fidelity; and incorporating realistic noise (e.g., Gaussian, dropout) into synthetic LiDAR to better simulate real-world sensor behavior and reduce the sim-to-real gap.

\section*{Acknowledgment}
The Experiments section of this work was funded by the German Federal Ministry for Economic Affairs and Climate Action (BMWK) under grant agreement 19I21039A. Financial support was also provided by Virtual Vehicle Research GmbH, the parent organization of SETLabs Research GmbH.
 
{\small
\bibliographystyle{IEEEtran}
\bibliography{biblio}

\begin{thebibliography}{10}
\providecommand{\url}[1]{#1}
\csname url@samestyle\endcsname
\providecommand{\newblock}{\relax}
\providecommand{\bibinfo}[2]{#2}
\providecommand{\BIBentrySTDinterwordspacing}{\spaceskip=0pt\relax}
\providecommand{\BIBentryALTinterwordstretchfactor}{4}
\providecommand{\BIBentryALTinterwordspacing}{\spaceskip=\fontdimen2\font plus
\BIBentryALTinterwordstretchfactor\fontdimen3\font minus \fontdimen4\font\relax}
\providecommand{\BIBforeignlanguage}[2]{{%
\expandafter\ifx\csname l@#1\endcsname\relax
\typeout{** WARNING: IEEEtran.bst: No hyphenation pattern has been}%
\typeout{** loaded for the language `#1'. Using the pattern for}%
\typeout{** the default language instead.}%
\else
\language=\csname l@#1\endcsname
\fi
#2}}
\providecommand{\BIBdecl}{\relax}
\BIBdecl

\bibitem{iec62267}
\emph{{IEC 62267: Railway Applications – Automated Urban Guided Transport (AUGT) – Safety Requirements}}, {International Electrotechnical Commission (IEC)} Std., 2009, accessed: 2025-04-29.

\bibitem{dosovitskiy2017carla}
A.~Dosovitskiy, G.~Ros, F.~Codevilla, A.~Lopez, and V.~Koltun, ``Carla: An open urban driving simulator,'' in \emph{Conference on robot learning}.\hskip 1em plus 0.5em minus 0.4em\relax PMLR, 2017, pp. 1--16.

\bibitem{nvidia_drivesim}
{NVIDIA Corporation}, ``Nvidia drive sim,'' \url{https://developer.nvidia.com/drive/drive-sim}, 2022.

\bibitem{SSDA3D_2023}
\BIBentryALTinterwordspacing
Y.~Wang, J.~Yin, W.~Li, P.~Frossard, R.~Yang, and J.~Shen, ``Ssda3d: semi-supervised domain adaptation for 3d object detection from point cloud,'' in \emph{Proceedings of the Thirty-Seventh AAAI Conference on Artificial Intelligence and Thirty-Fifth Conference on Innovative Applications of Artificial Intelligence and Thirteenth Symposium on Educational Advances in Artificial Intelligence}, ser. AAAI'23/IAAI'23/EAAI'23.\hskip 1em plus 0.5em minus 0.4em\relax AAAI Press, 2023. [Online]. Available: \url{https://doi.org/10.1609/aaai.v37i3.25370}
\BIBentrySTDinterwordspacing

\bibitem{shah2018airsim}
S.~Shah, D.~Dey, C.~Lovett, and A.~Kapoor, ``Airsim: High-fidelity visual and physical simulation for autonomous vehicles,'' in \emph{Field and Service Robotics: Results of the 11th International Conference}.\hskip 1em plus 0.5em minus 0.4em\relax Springer, 2018, pp. 621--635.

\bibitem{d2023trainsim}
G.~D’Amico, M.~Marinoni, F.~Nesti, G.~Rossolini, G.~Buttazzo, S.~Sabina, and G.~Lauro, ``Trainsim: A railway simulation framework for lidar and camera dataset generation,'' \emph{IEEE Transactions on Intelligent Transportation Systems}, vol.~24, no.~12, pp. 15\,006--15\,017, 2023.

\bibitem{de2023scenario}
J.~A.~I. de~Gordoa, S.~Garc{\'\i}a, I.~Urbieta, N.~Aranjuelo, M.~Nieto, D.~O. de~Eribe \emph{et~al.}, ``Scenario-based validation of automated train systems using a 3d virtual railway environment,'' in \emph{2023 IEEE 26th International Conference on Intelligent Transportation Systems (ITSC)}.\hskip 1em plus 0.5em minus 0.4em\relax IEEE, 2023, pp. 5072--5077.

\bibitem{toprak2020conditional}
T.~Toprak, B.~Belenlioglu, B.~Ayd{\i}n, C.~Guzelis, and M.~A. Selver, ``Conditional weighted ensemble of transferred models for camera based onboard pedestrian detection in railway driver support systems,'' \emph{IEEE Transactions on Vehicular Technology}, vol.~69, no.~5, pp. 5041--5054, 2020.

\bibitem{iglesias2024enhancing}
A.~Iglesias, J.~L. Apellaniz, N.~Aranjuelo, P.~Brandimarte, J.~Irastorza, and M.~Nieto, ``Enhancing safety in railway environments: Interpretable track and obstacle detection using on-board lidar,'' \emph{IEEE Transactions on Intelligent Transportation Systems (T-ITS)}, 2024, under review.

\bibitem{d2025syndra}
G.~D’Amico, F.~Nesti, G.~Rossolini, M.~Marinoni, S.~Sabina, and G.~Buttazzo, ``Syndra: Synthetic dataset for railway applications,'' in \emph{2025 IEEE/CVF Winter Conference on Applications of Computer Vision (WACV)}.\hskip 1em plus 0.5em minus 0.4em\relax IEEE, 2025, pp. 3437--3446.

\bibitem{geiger2012we}
A.~Geiger, P.~Lenz, and R.~Urtasun, ``Are we ready for autonomous driving? the kitti vision benchmark suite,'' in \emph{2012 IEEE conference on computer vision and pattern recognition}.\hskip 1em plus 0.5em minus 0.4em\relax IEEE, 2012, pp. 3354--3361.

\bibitem{behley2019semantickitti}
J.~Behley, M.~Garbade, A.~Milioto, J.~Quenzel, S.~Behnke, C.~Stachniss, and J.~Gall, ``Semantickitti: A dataset for semantic scene understanding of lidar sequences,'' in \emph{Proceedings of the IEEE/CVF international conference on computer vision}, 2019, pp. 9297--9307.

\bibitem{caesar2020nuscenes}
H.~Caesar, V.~Bankiti, A.~H. Lang, S.~Vora, V.~E. Liong, Q.~Xu, A.~Krishnan, Y.~Pan, G.~Baldan, and O.~Beijbom, ``nuscenes: A multimodal dataset for autonomous driving,'' in \emph{Proceedings of the IEEE/CVF conference on computer vision and pattern recognition}, 2020, pp. 11\,621--11\,631.

\bibitem{sun2020scalability}
P.~Sun, H.~Kretzschmar, X.~Dotiwalla, A.~Chouard, V.~Patnaik, P.~Tsui, J.~Guo, Y.~Zhou, Y.~Chai, B.~Caine \emph{et~al.}, ``Scalability in perception for autonomous driving: Waymo open dataset,'' in \emph{Proceedings of the IEEE/CVF conference on computer vision and pattern recognition}, 2020, pp. 2446--2454.

\bibitem{tagiew2025railgoerl24}
R.~Tagiew, I.~Wunderlich, M.~Sastuba, and S.~Seitz, ``Railgoerl24: G$\backslash$" orlitz rail test center cv dataset 2024,'' \emph{arXiv preprint arXiv:2504.00204}, 2025.

\bibitem{zendel2019railsem19}
O.~Zendel, M.~Murschitz, M.~Zeilinger, D.~Steininger, S.~Abbasi, and C.~Beleznai, ``Railsem19: A dataset for semantic rail scene understanding,'' in \emph{Proceedings of the IEEE/CVF Conference on Computer Vision and Pattern Recognition Workshops}, 2019, pp. 0--0.

\bibitem{cao2022effective}
Z.~Cao, Y.~Qin, Z.~Xie, Q.~Liu, E.~Zhang, Z.~Wu, and Z.~Yu, ``An effective railway intrusion detection method using dynamic intrusion region and lightweight neural network,'' \emph{Measurement}, vol. 191, p. 110564, 2022.

\bibitem{zouaoui2022railset}
A.~Zouaoui, A.~Mahtani, M.~A. Hadded, S.~Ambellouis, J.~Boonaert, and H.~Wannous, ``Railset: A unique dataset for railway anomaly detection,'' in \emph{2022 IEEE 5th International Conference on Image Processing Applications and Systems (IPAS)}.\hskip 1em plus 0.5em minus 0.4em\relax IEEE, 2022, pp. 1--6.

\bibitem{broekman2021railenv}
A.~Broekman and P.~J. Gr{\"a}be, ``Railenv-pasmvs: A perfectly accurate, synthetic, path-traced dataset featuring a virtual railway environment for multi-view stereopsis training and reconstruction applications,'' \emph{Data in Brief}, vol.~38, p. 107411, 2021.

\bibitem{neri20223d}
M.~Neri and F.~Battisti, ``3d object detection on synthetic point clouds for railway applications,'' in \emph{2022 10th European Workshop on Visual Information Processing (EUVIP)}.\hskip 1em plus 0.5em minus 0.4em\relax IEEE, 2022, pp. 1--6.

\bibitem{tagiew2023osdar23}
R.~Tagiew, P.~Klasek, R.~Tilly, M.~K{\"o}ppel, P.~Denzler, P.~Neumaier, T.~Klockau, M.~Boekhoff, and K.~Schwalbe, ``Osdar23: Open sensor data for rail 2023,'' in \emph{2023 8th International Conference on Robotics and Automation Engineering (ICRAE)}.\hskip 1em plus 0.5em minus 0.4em\relax IEEE, 2023, pp. 270--276.

\bibitem{Centerpoints}
T.~Yin, X.~Zhou, and P.~Kr{\"a}henb{\"u}hl, ``Center-based 3d object detection and tracking,'' \emph{CVPR}, 2021.

\bibitem{yang2021st3d}
J.~Yang, S.~Shi, Z.~Wang, H.~Li, and X.~Qi, ``St3d: Self-training for unsupervised domain adaptation on 3d object detection,'' in \emph{Proceedings of the IEEE/CVF conference on computer vision and pattern recognition}, 2021, pp. 10\,368--10\,378.

\bibitem{openpcdet2020}
O.~D. Team, ``Openpcdet: An open-source toolbox for 3d object detection from point clouds,'' \url{https://github.com/open-mmlab/OpenPCDet}, 2020.

\end{thebibliography}
}


\end{document}